\documentclass[11pt,a4paper]{article}
\usepackage[hyperref]{acl2019}
\usepackage{times}
\usepackage{latexsym}
\usepackage{graphicx}
\usepackage{booktabs}
\usepackage{multirow}
\usepackage{xspace}
\usepackage{url}
\aclfinalcopy
\def\aclpaperid{7}

\title{An Analysis of Emotion Communication Channels in Fan Fiction: \linebreak Towards Emotional Storytelling}

\author{
Evgeny Kim \and Roman Klinger\\
Institut f\"ur Maschinelle Sprachverarbeitung\\
University of Stuttgart\\
Pfaffenwaldring 5b, 70569 Stuttgart, Germany\\
\texttt{\{evgeny.kim,roman.klinger\}@ims.uni-stuttgart.de}
}

\date{}

\begin{document}
\maketitle
\begin{abstract}
  Centrality of emotion for the stories told by humans is underpinned
  by numerous studies in literature and psychology. The research in
  automatic storytelling has recently turned towards emotional
  storytelling, in which characters' emotions play an important role
  in the plot development
  \cite{theune2004emotional,y2007employing,mendez2016use}. However,
  these studies mainly use emotion to generate propositional
  statements in the form ``\textit{A} feels affection towards
  \textit{B}" or ``\textit{A} confronts \textit{B}". At the same time,
  emotional behavior does not boil down to such propositional
  descriptions, as humans display complex and highly variable patterns
  in communicating their emotions, both verbally and non-verbally. In
  this paper, we analyze how emotions are expressed non-verbally in a
  corpus of fan fiction short stories. Our analysis shows that stories
  written by humans convey character emotions along various non-verbal
  channels. We find that some non-verbal channels, such as facial
  expressions and voice characteristics of the characters, are more
  strongly associated with \textit{joy}, while gestures and body
  postures are more likely to occur with \textit{trust}. Based on our
  analysis, we argue that automatic storytelling systems should take
  variability of emotion into account when generating descriptions of
  characters' emotions.
\end{abstract}

\section{Introduction and Motivation}
\label{intro}
As humans, we make sense of our experiences through stories
\cite{mckee2003story}. A key component of any captivating story is a
character \cite{kress2005characters} and a key component of every
character is emotion, as ``without emotion a character's personal
journey is pointless'' \cite[p.\ 1]{emothesaurus}. Numerous works
pinpoint the central role of emotions in storytelling
\cite{hogan2015literature,johnson2016emotion,ingermanson2009writing},
as well as story comprehension and evaluation
\cite{Komeda2005,van1997characters,mori2019}.

\begin{figure}[b]
\centering
\includegraphics[width=\linewidth]{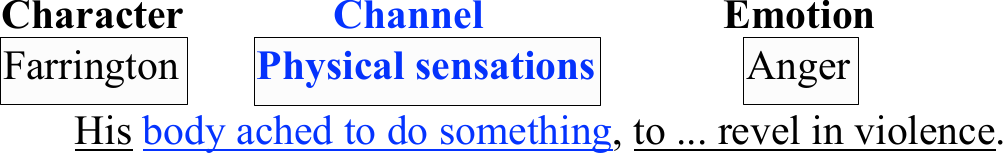}
\caption{Example of the emotion expressed using non-verbal
  communication channel. The annotation of \textit{character}
  and \textit{emotion} are available in the dataset by
  \newcite{Kim2019}. \textit{Channel} annotation (in blue) is
  an extension to the original dataset.}
\label{annotation}
\end{figure}

Emotion analysis and automatic recognition in text is mostly
channel-agnostic, \textit{i.e.}, does not consider along which
non-verbal communication channels (face, voice, etc.) emotions are
expressed. However, we know that the same emotions can be expressed
non-verbally in a variety of ways \cite[p.\ 11]{barrett2017emotions},
for example, through internal feelings of the character, as shown in
Figure \ref{annotation}. We argue that automatic storytelling systems
should take this information into account, as versatility of the
emotion description is a prerequisite for engaging and believable
storytelling \cite[p.\ 3]{emothesaurus}.

There is a growing body of literature in the field of natural language
generation that uses emotions as a key component for automatic plot
construction \cite{theune2004emotional,y2007employing,mendez2016use}
and characterization of virtual agents
\cite{imbert2005emotional,dias2011agents}. However, these and other
studies put an emphasis on emotion \textit{per se} (``\textit{A} feels
affection towards \textit{B}''), or on the social behavior of
characters ``\textit{A} confronts \textit{B}'') making little or no
reference to how characters express emotions, both verbally and
non-verbally.

In this paper, we aim at closing this gap by analyzing how characters
express their emotions using non-verbal communication
signals. Specifically, we analyze how eight emotions (\textit{joy},
\textit{sadness}, \textit{anger}, \textit{fear}, \textit{trust},
\textit{disgust}, \textit{surprise}, and \textit{anticipation})
defined by \newcite{Plutchik2001} are expressed along the following
channels introduced by \newcite{VANMEEL1995159}: 1)~physical
appearance, 2)~facial expressions, 3)~gaze, looking behavior, 4)~arm
and hand gesture, 5)~movements of body as a whole, 6)~characteristics
of voice, 7)~spatial relations, and 8)~physical make-up.

This paper is an extension to our previous study \cite{Kim2019}, in
which we presented a corpus of emotion relations between characters in
fan fiction short stories. We post-annotate the corpus with non-verbal
expressions of emotions and analyze two scenarios of non-verbal
emotion expression: when the feeler of an emotion is alone, and when a
communication partner, who also plays a role in the development of
emotion, is present. In our analysis, we look into the emotions
associated with each non-verbal behavior, mapping emotions to
non-verbal expressions they frequently occur with.

Our contributions are therefore the following: 1)~we propose that
natural language generation systems describing emotions should take
into account how emotions are expressed non-verbally, 2)~we extend the
annotations presented in \newcite{Kim2019} and quantitatively analyze
the data, 3)~we show that facial expressions, voice, eyes and body
movements are the top three channels among which the emotion is
expressed, 4)~based on the data, we show that some emotions are more
likely to be expressed via a certain channel, and this channel is also
influenced by the presence or non-presence of a communication partner.

Our corpus is available at \url{https://www.ims.uni-stuttgart.de/data/emotion}.

\section{Related Work}
Emotion analysis has received great attention in natural language
processing
\cite[\textit{i.a.}]{mohammad-bravo-marquez-2017-wassa,mohammad-etal-2018-semeval,Klinger2018x,felbo-etal-2017-using,abdul-mageed-ungar-2017-emonet,zhou-wang-2018-mojitalk,gui-etal-2017-question}. Most
existing studies on the topic cast the problem of emotion analysis as
a classification task, by classifying documents (\textit{e.g.}, social
media posts) into a set of predefined emotion classes. Emotion classes
used for the classification are usually based on discrete categories
of \newcite{Ekman1970} or \newcite{Plutchik2001} (\textit{cf.}\
\newcite{Alm:2005}, \newcite{10.1007/978-3-642-37256-8_11},
\newcite{mohammad2012once}). Fewer studies address emotion recognition
using a dimensional emotion representation (\textit{cf.}\
\newcite{buechel-hahn-2017-emobank,preotiuc-pietro-etal-2016-modelling}). Such
representation is based on the valence-arousal emotion model
\cite{russelcircumplex}, which can be helpful to account for
subjective emotional states that do not fit into discrete categories.

Early attempts to computationally model emotions in literary texts
date back to the 1980s and are presented in the works by
\newcite{anderson1982computer,anderson1986modeling}, who build a
computational model of affect in text tracking how emotions develop in
a literary narrative.

More recent studies in the field of digital humanities approach
emotion analysis from various angles and for a wide range of
goals. Some studies use emotions as feature input for genre
classification
\cite{samothrakis2015emotional,krahmerSentiment2018,yubei2008,kim-etal-2017-investigating},
story clustering \cite{reagan2016emotional}, mapping emotions to
geographical locations in literature \cite{heuserlondon2016}, and
construction of social networks of characters
\cite{nalisnick2013extracting,Jhavar:2018:EME:3184558.3186989}. Other
studies use emotion analysis as a starting point for stylometry
\cite{koolen2018}, inferring psychological characters' traits
\cite{egloff2018}, and analysis of the causes of emotions in
literature \cite{kim-klinger-2018-feels,Kim2019}.

To the best of our knowledge, there is no previous research that
addresses the question of how emotions are expressed non-verbally. The
only work that we are aware of is a literary study by
\newcite{VANMEEL1995159}, who proposes several non-verbal
communication channels for emotions and performs a manual analysis
on a set of several books. He finds that voice is the most frequently
used category, followed by facial expressions, arm and hand gestures
and bodily postures. Van Meel explains the dominancy of voice by the
predominant role that speech plays in novels. However, van Meel does
not link the non-verbal channels to any specific emotions. In this
paper, we extend his analysis by mapping the non-verbal channels to a
set of specific emotions felt by the characters.

\begin{table*}[tbp]
\centering
\begin{tabular}{lrrrrrrrrrrrrrrrrrrrr}
\toprule
Emotion && \rotatebox{90}{Face} && \rotatebox{90}{Body} && \rotatebox{90}{Appear.} && \rotatebox{90}{Look.} && \rotatebox{90}{Voice} && \rotatebox{90}{Gesture} && \rotatebox{90}{Sptrel.} && \rotatebox{90}{Sensations} && \rotatebox{90}{No channel} &&\rotatebox{90}{\textbf{Total}} \\
\cmidrule{1-1}\cmidrule{3-3}\cmidrule{5-5}\cmidrule{7-7}\cmidrule{9-9}\cmidrule{11-11}\cmidrule{13-13}\cmidrule{15-15}\cmidrule{17-17}\cmidrule{19-19}\cmidrule{21-21}
anger && 23 && 20 && 5 && 38 && 51 && 7 && 0 && 4 && 163 && 311\\
anticipation && 4 && 14 && 0 && 17 && 4 && 2 && 7 && 6 && 267 && 321\\
disgust && 3 && 6 && 1 && 3 && 0 && 0 && 0 && 1 && 149 && 163\\
fear && 4 && 28 && 13 && 16 && 8 && 1 && 0 && 25 && 124 && 219\\
joy && 76 && 26 && 7 && 12 && 52 && 19 && 5 && 33 && 268 && 498\\
sadness && 3 && 5 && 4 && 4 && 2 && 0 && 3 && 7 && 81 && 109\\
surprise && 10 && 5 && 3 && 13 && 1 && 0 && 0 && 2 && 118 && 152\\
trust && 4 && 15 && 1 && 4 && 1 && 21 && 3 && 0 && 144 && 193\\
\cmidrule{1-1}\cmidrule{3-3}\cmidrule{5-5}\cmidrule{7-7}\cmidrule{9-9}\cmidrule{11-11}\cmidrule{13-13}\cmidrule{15-15}\cmidrule{17-17}\cmidrule{19-19}\cmidrule{21-21}
\textbf{Total} && 127 && 119 && 34 &&  107 && 119 && 50 && 18 && 78 && 1314 && 1966 \\
\bottomrule
\end{tabular}
\caption{Counts of emotion and expression-channel pairs. \textit{No channel} means that instance contains no reference to how emotion is expressed non-verbally.}
\label{modality_channels}
\end{table*}

\section{Corpus Creation}
We post-annotate our dataset of emotion relations between characters
in fan fiction \cite{Kim2019} with non-verbal communication channels
of emotion expressions. The dataset includes complete annotations of
19 fan fiction short stories and of one short story by James
Joyce. The emotion relation is characterized by a triple
$(C_{\textrm{exp}},e,C_{\textrm{cause}})$, in which the character
$C_{\textrm{exp}}$ feels the emotion $e$. The character
$C_{\textrm{cause}}$ (to which we refer as a ``communication partner")
is part of an event which triggers the emotion $e$. The emotion
categorization presented in the dataset follows Plutchik's
(\citeyear{Plutchik2001}) classification, namely \textit{joy},
\textit{sadness}, \textit{anger}, \textit{fear}, \textit{trust},
\textit{disgust}, \textit{surprise}, and \textit{anticipation}.

Given an annotation of a character with an emotion, we annotate
non-verbal channels of emotion expressions following the
classification proposed by \newcite{VANMEEL1995159}, who defines the
following eight categories: 1) physical appearance, 2) facial
expressions, 3) gaze, looking behavior, 4) arm and hand gesture, 5)
movements of body as a whole, 6) characteristics of voice, 7) spatial
relations (references to personal space), and 8) physical make-up. To
clarify the category of \textit{physical make-up}, we redefine it
under the name of \textit{physical sensations}, \textit{i.e.},
references to one's internal physiological signals perceived by the
feeler of the emotion.

The task is exemplified in Figure \ref{annotation}. Labels for emotion
(\textit{Anger}) and character (\textit{Farrington}) are given.
\textit{Physical sensation} is an example of a channel annotation we
use to extend the original dataset.

The annotations were done by three graduate students in our
computational linguistics department within a one-month period. The
annotators were asked to read each datapoint in the dataset and decide
if the emotion expressed by the feeler (\textit{character}) has an
associated non-verbal channel of expression. If so, the annotators
were instructed to mark the corresponding textual span and select a
channel label from the list of non-verbal communication channels given
above.

\begin{table}
\centering
\begin{tabular}{lccc} 
\toprule
& \multicolumn{1}{l}{a1--a2} & \multicolumn{1}{l}{a1--a3} & \multicolumn{1}{l}{a2--a3} \\
\cmidrule(lr){2-2}\cmidrule(lr){3-3}\cmidrule(lr){4-4}
Span & 31 & 29 & 45 \\
Sentence & 49 & 45 & 59 \\
\bottomrule
\end{tabular}%
\caption{F$_1$ scores in \,\% for agreement between annotators on a span level. a1, a2, and a3 are different annotators. Span: label of channel and offsets are considered. Sentence: only label of the channel in the sentence is considered.}
\label{iaafanfic}%
\end{table}

The results of inter-annotator agreement (IAA) are presented in Table
\ref{iaafanfic}. We measure agreement along two dimensions: 1)~span,
where we measure how well two people agree on the label of a
non-verbal emotion expression, as well as on the exact textual
position of this expression, and 2)~sentence, where we measure how
well two people agree on the label of non-verbal emotion expression in
a given sentence (\textit{i.e.}, the exact positions of the channel
are not taken into account). The agreement is measured using the F$_1$
measure, where we assume that annotations by one person are true, and
annotations by another person are predicted. As one can see, the
agreement scores for spans (\textit{i.e.}, channel label and exact
textual positions) are relatively low (lowest 29\%, highest 45\% F$_1$
respectively). The IAA scores on a sentence level are higher (lowest
agreement is 45\%, highest 59\% F$_1$ respectively), as we only
consider the label of the non-verbal channel in a sentence without
looking into the exact textual positions of the annotations.

\begin{table*}[t]
	\centering
	\begin{tabular}{llllll}
\toprule
Channel && Emotion && Examples \\
\cmidrule{1-1}\cmidrule{3-3}\cmidrule{5-5}
Facial expressions && anger && rolled his eyes  \\
&& fear && smiled nervously \\
\cmidrule{1-1}\cmidrule{3-3}\cmidrule{5-5}
Body movements  && anger && stormed back out \\
&& trust  && slumped his shoulders \\
\cmidrule{1-1}\cmidrule{3-3}\cmidrule{5-5}
Physical appearance&& fear && blushed crimson red \\
\cmidrule{1-1}\cmidrule{3-3}\cmidrule{5-5}
Looking behavior && fear && averted her eyes \\
&& anticipation && pause to look back \\
\cmidrule{1-1}\cmidrule{3-3}\cmidrule{5-5}
Voice && joy && purred \\
&& fear && voice getting smaller and smaller \\
\cmidrule{1-1}\cmidrule{3-3}\cmidrule{5-5}
Arm and hand gestures && trust && opened her arms \\
\cmidrule{1-1}\cmidrule{3-3}\cmidrule{5-5}
Spatial relations && joy && leaping into her arms \\
&& trust && pulled him closer to his chest \\
\cmidrule{1-1}\cmidrule{3-3}\cmidrule{5-5}
Physical sensations && joy && tingling all over \\
&& fear && hear in his throat \\
\bottomrule
\end{tabular}
\caption{Textual examples of non-verbal emotion expressions.}
\label{Examples}
\end{table*}

\section{Analysis}
\label{analysis}

Table \ref{modality_channels} summarizes the results of the annotation
of non-verbal channels of emotion expressions, Table \ref{Examples}
gives examples of these expressions in the dataset.

In total, there are 652 annotations of non-verbal emotion
expressions. By absolute counts, facial expressions (\textit{Face},
127 occurrences), body movements (\textit{Body}, 119), voice
(\textit{Voice}, 199), and looking behavior (\textit{Look.}, 107) have
the highest number of annotations. Spatial relations
(\textit{Sptrel.}, 78) and physical appearance (\textit{Appear.}, 34)
have the lowest number of annotations.

\subsection{Emotion-Channel associations}

\begin{figure}
  \includegraphics[width=\linewidth]{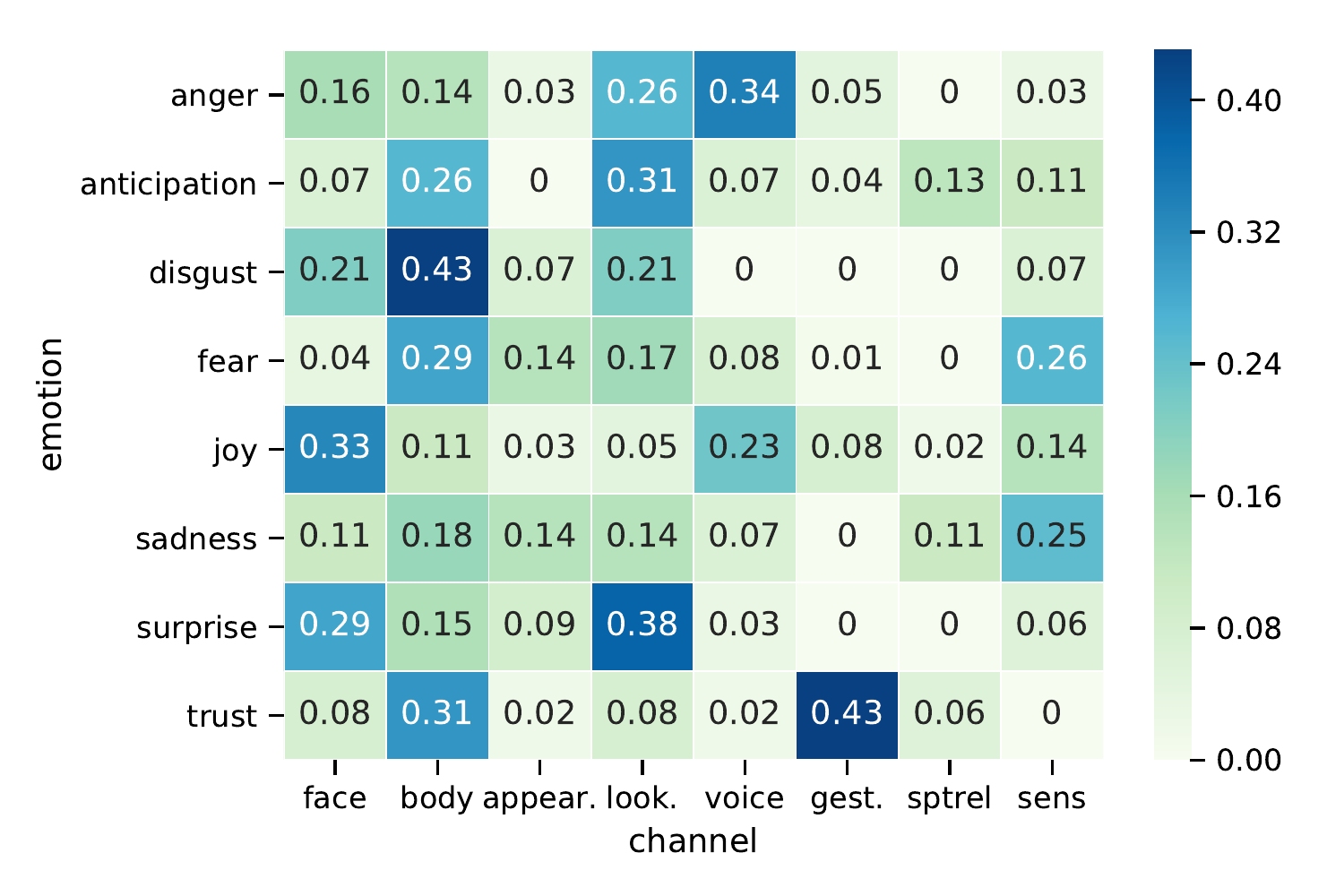}
  \caption{Distribution of non-verbal channels with all
    emotions. Darker color indicates higher value. Values in the cells
    are percentage ratios. Each cell is normalized by the row sum of
    absolute frequencies.}
  \label{heatmap1}
\end{figure}

We start our analysis by looking into the emotion-channel
associations. Namely, we analyze which non-verbal channels are
associated with which emotions. To that end, we plot a heatmap of
the emotion--non-verbal-channel matrix. The value of each cell in the
heatmap is normalized by the row sum (\textit{i.e.}, total counts of
channel annotations) and represents the likelihood of emotion-channel
association in the dataset, for each emotion, respectively.

As Figure \ref{heatmap1} shows, \textit{anger} is more likely to be
expressed with \textit{voice}, while \textit{joy} is more likely to be
expressed with \textit{face}. In contrast to all other emotions,
\textit{sadness} is more likely to be experienced internally
(\textit{sens.} column in Figure \ref{heatmap1}) by the feeler, as
opposed to being communicated non-verbally. Some channels and emotions
show no association, such as \textit{gestures} (\textit{gest.}) and
\textit{disgust} or \textit{spatial relations} (\textit{sptrel.}) and
\textit{anger}. Given the relatively small size of the dataset, we do
not argue that these associations are not possible in principle. For
example, \textit{fear} and \textit{spatial relations} have zero
association in our analysis, however, it is likely that somebody
expresses this emotion by moving away subconsciously (increasing
personal space) from the source of danger. At the same time,
\textit{fear} is most strongly associated with \textit{body movements}
as a whole, which can be difficult to distinguish from spatial
relations. However, we believe that these associations still reflect
the general trend: emotions that require immediate actions and serve
evolutionary survival function, such as anger, disgust, and fear, are
more likely to be expressed with actions. For example, anger is an
unpleasant emotion that often occurs as a response to an appraisal of
a blocked goal \cite{harmonjonesAnger}, which can be resolved by using
voice characteristics (commanding or shouting at someone who prevents
you from achieving your goal).
 
\begin{figure}
 \centering
 \includegraphics[width=\linewidth,height=4cm]{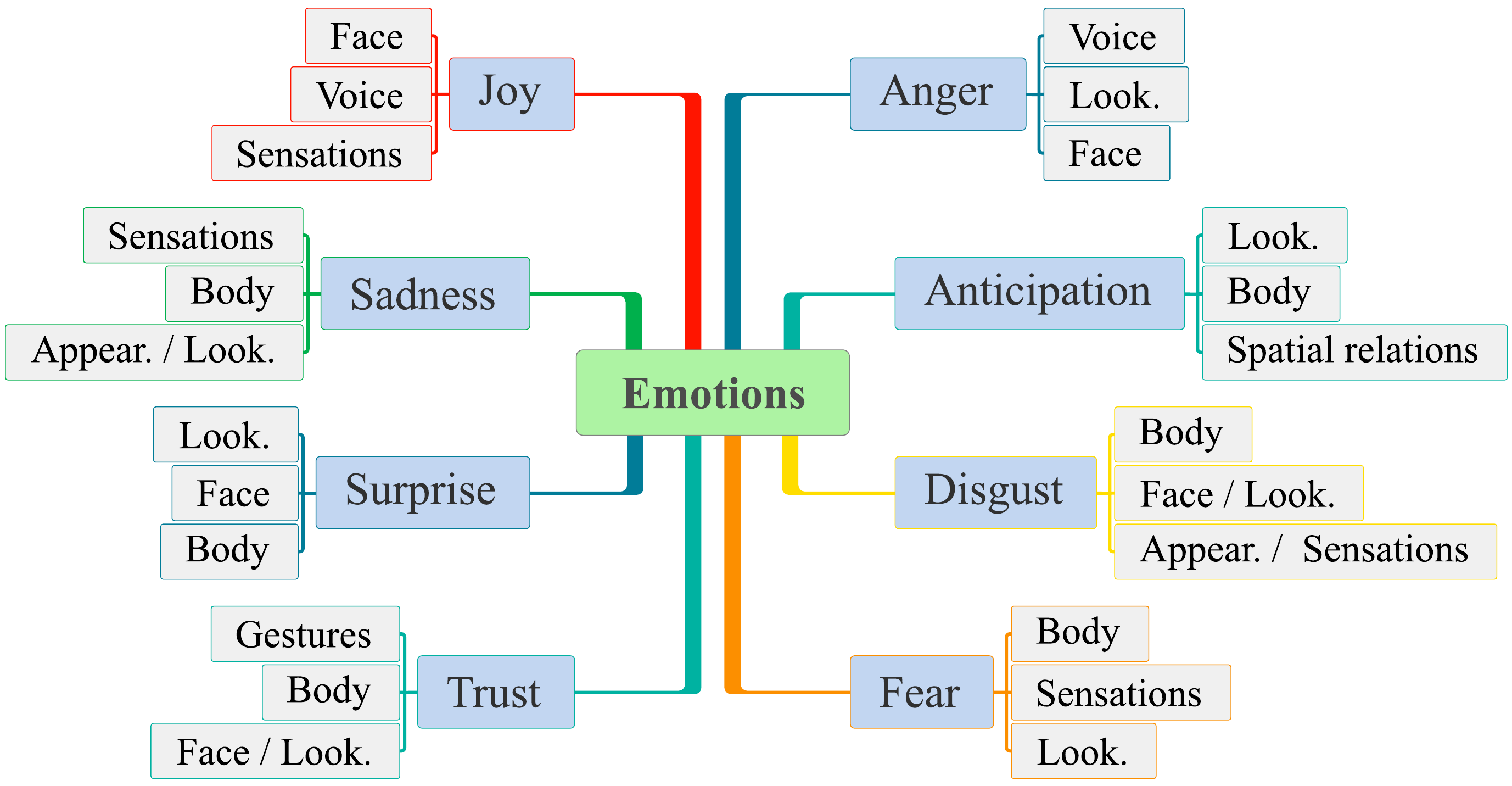}
 \caption{Emotion-channel map. Each branch is an emotion, whose branches are the top three non-verbal channels associated with the emotion.}
 \label{emotionmap}
\end{figure}
 
Overall, we observe that \textit{face}, \textit{look.},
\textit{voice}, and \textit{body} channels are predominantly used with
all emotions. We visualize the strongest associations of emotions and
non-verbal channels in Figure \ref{emotionmap}. For each emotion, the
figure shows the top three (in a descending order) strongly associated
non-verbal channels. As one can see, the branches are dominated by
\textit{face}, \textit{look.}, \textit{voice}, and \textit{body}
channels. The only exception is \textit{trust}, for which the
predominant way to express emotions non-verbally is through
\textit{gestures}, and \textit{sadness}, which is predominantly felt
``internally'' (\textit{sensations}).

\subsection{Presence of a communication partner}
\begin{figure}[t]
  \includegraphics[width=\linewidth]{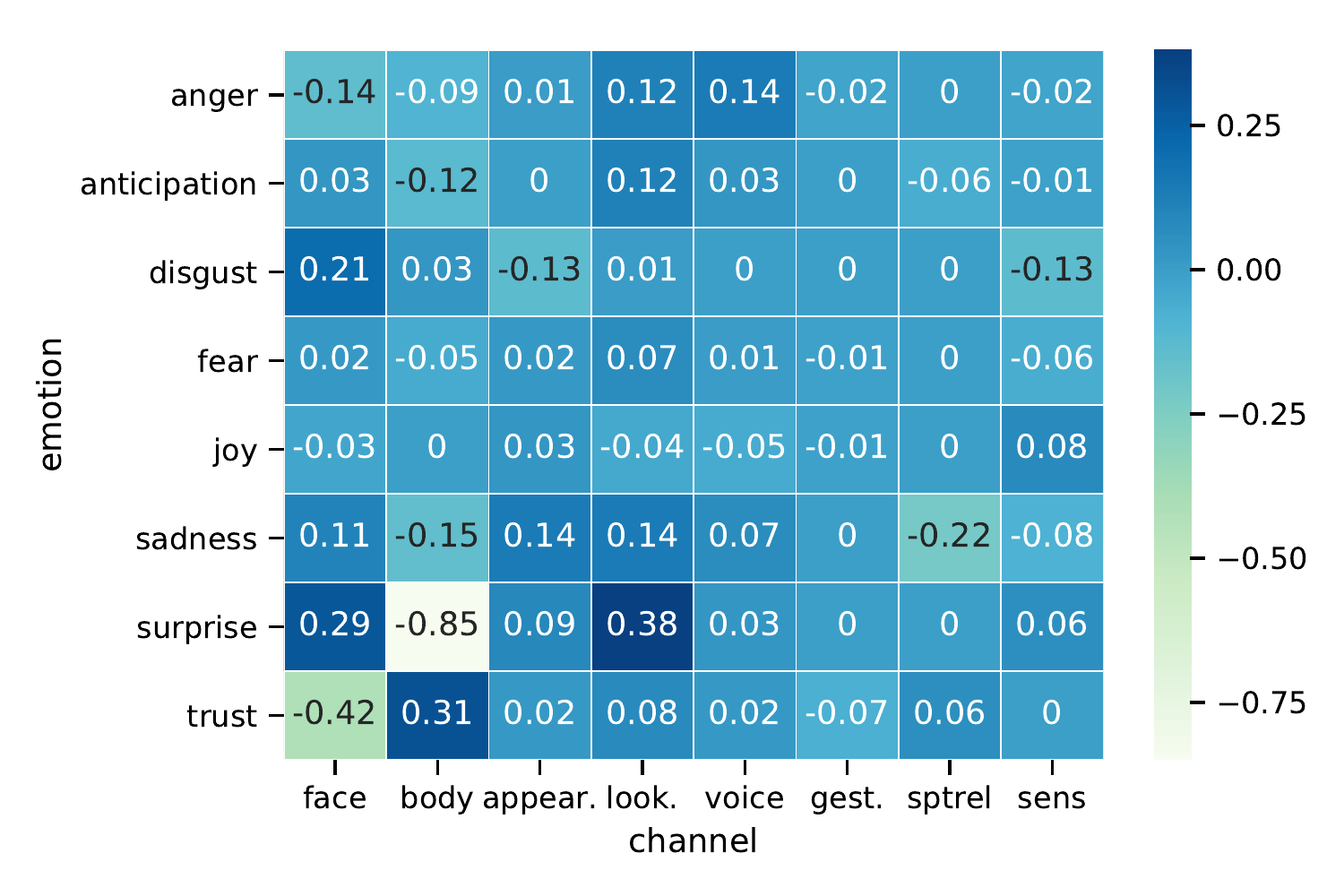}
  \caption{The difference between situations, in which a character
    feels an emotion and the communication partner is present, and
    situations in which the communication partner is not present
    (normalized by row sum). Darker color/higher values indicates that
    the channel is more likely to be used when there is a
    communication partner.}
  \label{heatmap2}
\end{figure}
\begin{figure*}
  \includegraphics[width=\linewidth]{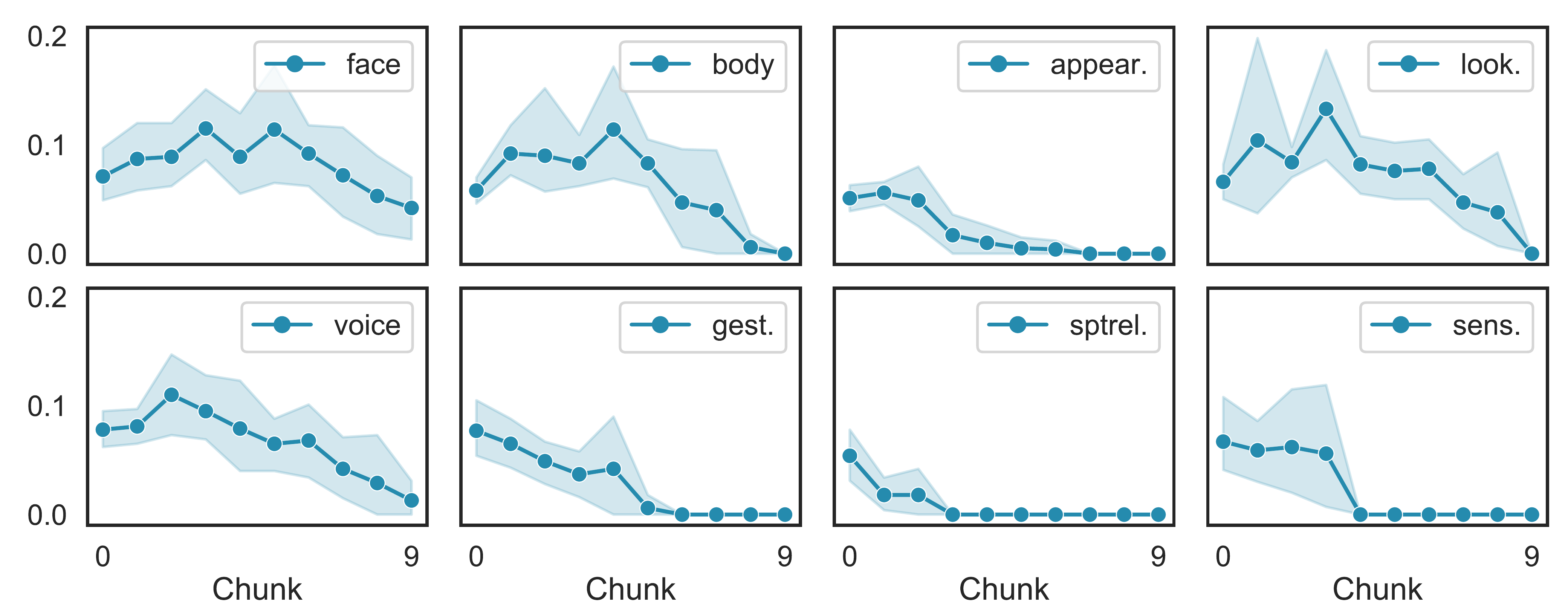}
  \caption{Distribution of non-verbal emotion expressions in the narrative. Markers on the plot lines indicate the text chunk. The plots are given for ten chunks. Light area around the solid line indicates confidence interval of 95\%. \textit{y-}axis shows percentage.}
  \label{timeline}
\end{figure*}
The original dataset contains information whether an emotion of the
character is evoked by another character (communication partner). We
use this information to understand how the presence of a communication
partner affects the choice of a non-verbal channel.

To that end, we plot a heatmap (Figure \ref{heatmap2}) from the delta
values between situations, in which the communication partner is
present, and situations in which the communication partner is not
present. As it can be seen from Figure \ref{heatmap2}, \textit{trust}
is more strongly associated with \textit{body movements} when a
communication partner is present. \textit{Sadness}, which is more
likely to associate with inner physical sensations in the feeler, is
expressed through the \textit{physical appearance} and \textit{looking
  behavior} when the communication partner is present. Likewise,
\textit{disgust} is more likely to be expressed with \textit{facial
  expressions}, and \textit{anticipation} is more likely to be
expressed with \textit{looking behavior}.

Again, we observe that \textit{body}, \textit{voice}, \textit{face},
and \textit{look.} channels are the predominant non-verbal
communication channels for emotion expressions.

\subsection{Timeline analysis}
To understand if there are patterns in the frequency of use of
non-verbal channels in a narrative, we perform an additional analysis.

For this analysis, we split each short story in the dataset in ten
equally sized chunks and get frequencies of each non-verbal channel,
which are then plotted as time-series with confidence intervals
(Figure \ref{timeline}). The averaged values for each channel are
plotted as a dark line with circular markers. The lighter area around
the main line represents confidence intervals of 95\%, which are
calculated using bootstrap resampling.

We observe the general tendency of all non-verbal channels to vanish
towards the end of the story. The only exception is \textit{Facial
  expressions}, which after peaking in the middle of a story reverts
to the mean. Overall, we find no consistent pattern in the use of
non-verbal channels from beginning to an end of a story.

\section{Discussion and Conclusion}
The results of the analysis presented in Section \ref{analysis} show
that emotions are expressed in a variety of ways through different
non-verbal channels. However, the preferred communication channel
depends on whether a communication partner is present or not. Some
channels are used predominantly only when the feeler communicates her
emotion to another character, other channels can be used in any
situation.

\textit{Sadness} stands out from other emotions in a sense that it is
predominantly not expressed using any external channels of non-verbal
communication. In other words, it is more common for the characters in
the annotated dataset to go through sadness ``alone" and feel it ``in
the body", rather than show it to the outer world. However, when
another character (communication partner) is the reason of sadness
experienced by the feeler, he or she will most likely use eyes and
overall behavior to show this emotion.

In this paper, we showed that in human-written stories, emotions are
not only expressed as propositions in the form of ``\textit{A} feels
affection towards \textit{B}" or ``\textit{A} confronts
\textit{B}''. As Table \ref{Examples} shows, often there is no direct
mention of the feelings \textit{A} holds towards \textit{B} (``rolled
his eyes", ``purred"). It is, therefore, important, that this
observation finds its place in automatic storytelling systems. Some
attempts have been done in natural language generation towards
controllable story generation
\cite{peng-etal-2018-towards,tambwekar2018controllable}. We propose
that emotion expression should be one of the controllable parameters
in automatic storytellers. As more and more language generation
systems have started using emotion as one of the central components
for plot development and characterization of characters, there will be
a need for a more versatile and subtle description of emotions, which
is realized not only through propositional statements. In the end, no
single instance of same emotion is expressed in the same way
\cite{barrett2017emotions}, and emotion-aware storytelling systems
should take this information into account when generating emotional
profiles of characters.

\section{Future Work}
This paper proposes one approach to non-verbal emotion description
that relies on a rigid ontology of emotion classes. However, it might
be reasonable to make use of unsupervised clustering of non-verbal
descriptions to overcome the limitations of using a relatively small
number of coarse emotion categories for the description of character
emotions. Once clustered, such descriptions could be incorporated in
the generated text (\textit{e.g.}, a plot summary) and would elaborate
all the simplistic descriptions of character emotions.

Other research directions seems feasible too. For example, the
annotations, which we presented in this paper, can be used for
building and training a model that automatically recognizes non-verbal
channels of emotion expressions. This might, in a multi-task learning
setting, improve emotion classification. The data we provide could
also be used as a starting point for terminology construction, namely
bootstrapping a lexicon of emotion communications with different
channels. Finally, our work can serve as a foundation for the
development of an automatic storytelling system that takes advantage
of such resources.

\section*{Acknowledgements}
This research has been conducted within the CRETA project
(\url{http://www.creta.uni-stuttgart.de/}) which is funded by the
German Ministry for Education and Research (BMBF) and partially funded
by the German Research Council (DFG), projects SEAT (Structured
Multi-Domain Emotion Analysis from Text, KL 2869/1-1).

\end{document}